\documentclass[11pt]{article}

\usepackage[utf8]{inputenc}
\usepackage[T1]{fontenc}
\usepackage{mathpazo}               
\usepackage[scaled=0.85]{berasans} 
\usepackage[scaled=0.85]{beramono} 
\usepackage{microtype}             

\usepackage[margin=1in]{geometry}
\usepackage{setspace}
\usepackage{parskip}              

\usepackage{graphicx}
\usepackage{amsmath}
\usepackage{amssymb}
\usepackage{booktabs}
\usepackage{tabularx}
\usepackage{array}
\usepackage{enumitem}
\usepackage[font=small,labelfont=bf]{caption}

\usepackage{xcolor}
\usepackage[colorlinks,citecolor=blue!60!black,linkcolor=blue!60!black,urlcolor=blue!60!black]{hyperref}
\usepackage[round]{natbib}

\newcolumntype{Y}{>{\raggedright\arraybackslash}X}
\newcommand{\dhh}{\Delta h}
\newcommand{\WU}{W_U}

\title{Contrastive Projection: Reading Transformer Internals\\ by Differencing Logit Lenses}
\author{Olli Tuomi\\ Evident Solutions Oy}
\date{}

\begin{document}
\maketitle

\begin{abstract}
Reading a transformer's internal states in token space is easy to do and hard
to trust: a logit lens on a single hidden state is dominated, at intermediate
layers, by the generic tokens the model would predict for almost any input. We
read the difference instead. Subtracting two closely matched prompts' hidden
states and projecting through the unembedding cancels the shared component and
surfaces what separates them, an operation equivalent to reading a RepE/ActAdd 
steering vector through a logit lens. Built into a training-free tracer that reads at
every position, sub-layer, and head and averages over designed baselines, it
traces a compound-noun MLP$\rightarrow$attention chain in Phi-2, confirmed there
by activation patching, with the same distinction recovered across three
architectures by readout and probe rather than by patching; it reads what
retrieval surfaces for real versus fictional entities, and reads metaphor as a
set of domain-to-domain mappings rather than a single figurativity feature. A
cross-seed control marks the boundary: across five networks differing only in
initialization, the same distinction surfaces as almost entirely different
tokens (top-10 overlap $0.08$). What a computation looks like in token space is
network-specific; the distinction it draws is not.
\end{abstract}

\section{Introduction}

A transformer processing ``The hot dog was'' predicts continuations about food.
The same model given ``The cold dog was'' predicts continuations about an
animal. The logit lens~\citep{nostalgebraist2020} projects each hidden state through
$\WU$. But the single-input projection is dominated by strong shared signals.
At intermediate layers it reads the same high-frequency function words for both
inputs (``not, no, made, more''), not the food or animal content
(Table~\ref{tab:hotdog}, first column).

Subtracting one hidden state from the other before projecting through $\WU$
cancels the shared signals and reads what differs. The hot dog$-$cold dog difference
reads food vocabulary from layer 8 on (``fried, crispy, delicious, flavor''),
absent from either constituent's logit-lens top-20 (0--1/5 overlap through
L24). The same contrast read the other way, cold$-$hot, surfaces animal
vocabulary (``grooming, Paw, Bark, breeds''), the pet reading that the food
direction cancels. The difference is signed: each direction reports what one
side carries that the other does not.

Two closely matched inputs share most of their computation, so what survives
the subtraction is the component where the model treats them differently.

\begin{table}[htbp]
\centering
\small
\begin{tabular}{@{}rlll@{}}
\toprule
L & Raw (both) & hot $-$ cold & cold $-$ hot \\
\midrule
8  & nt, no, party   & fried, hors, crispy     & vet, Pent, ogie \\
22 & not, made, more & tast, delicious, flavor & grooming, euth, Paw \\
24 & not, more, so   & delicious, tast, flavor & grooming, Paw, Bark \\
\bottomrule
\end{tabular}
\caption{The opening contrast, read at the final token of ``The hot dog was''
versus ``The cold dog was'' (Phi-2). The raw logit lens of either input reads
high-frequency function words at every layer; the hot$-$cold difference reads
food vocabulary (legible from L8), the cold$-$hot difference reads animal
vocabulary (sharpening by L20). Verbatim top-3; some entries are sub-word
fragments (``nt,'' ``hors,'' ``euth'') or noise (``Pent, ogie'').}
\label{tab:hotdog}
\end{table}

Every contrastive reading is pair-specific. It reports what separates one
particular pair of prompts. A tight pair that varies one thing reads more
cleanly than a loose pair that varies several. Contrast design steers the
reading: what it surfaces is chosen, not discovered. This is a mechanistic form
of contrastive explanation, where one explains why one case rather than a
chosen foil, and the foil fixes the answer \citep{vanfraassen1980,lipton1990}.

The primitive is not new. Differencing two prompts' hidden states and
projecting through $\WU$ is the arithmetic of a Representation Engineering (RepE)
or Activation Addition (ActAdd) steering vector \citep{zou2023,turner2023} read
through a logit lens \citep{nostalgebraist2020}.
\citet{du2026} already used this exact operation. They decoded an activation
difference through the logit lens to trace control signals for reflection in
R1-style reasoning models. Our contribution is what this readout becomes as a
systematic tracer. Read at every position and sub-layer, and averaged over
designed baselines, it locates where an axis of variation first becomes legible
in token space, which sub-layer writes it, and which head carries it. We add
three techniques and a set of applications:

\begin{enumerate}[leftmargin=*]
\item \textbf{Systematic trajectory reading.} Per-position tracing locates where a
   distinction first appears and how it flows between positions. Per-head
   decomposition identifies which attention heads carry it. Layer-by-layer
   readout tracks how the content changes from early detection to final
   prediction.

\item \textbf{Multi-contrast triangulation.} A single pair reports every way the
   two prompts differ, not only the intended axis, so its readout can be hard
   to read. Contrast the target against several baselines that share the
   intended axis but differ on the incidental ones, then average. The
   incidental axes cancel and the shared one remains (\S\ref{sec:triangulation}).

\item \textbf{Contrast design.} What the readout surfaces depends on how the pair
   is built. We give rules for matched contrasts: a shared \emph{preamble} to
   avoid the massive-activation first token, a matched current token at the read
   position so the difference is content and not token identity, and a matched
   predicted next token to expose mid- and late-layer computation
   (\S\ref{sec:design}).
\end{enumerate}

We apply the method to Phi-2 (2.7B). We re-run the compound-noun circuit across
three architectures (Phi-2, Pythia-1.4B, and Qwen2.5-1.5B) to separate what
generalizes from what is model-specific (\S\ref{sec:crossmodel}). The cases
shown are selected from several hundred readings. We report the most
illustrative.

One caveat frames everything that follows. The specific tokens we read out
(``fried,'' ``Nepal, Tibet,'' ``delicious'') are illustrative, not a stable
code. A cross-seed control (\S\ref{sec:readout}) shows that five networks
differing only in initialization surface the same distinction as almost entirely
different tokens (top-10 overlap $0.08$), while the distinction itself holds in
every seed. Throughout, read a token as evidence of what a contrast separates in
one network, not as a canonical vocabulary.

\section{Method}

\subsection{Contrastive projection}

Given two inputs $c$ and $k$:

\begin{enumerate}[leftmargin=*]
\item Run both and extract hidden states at every layer at the read position:
   $h_c[L]$ and $h_k[L]$ for $L = 0, \dots, N$.
\item Compute $\dhh[L] = h_c[L] - h_k[L]$.
\item Project: $\Delta\text{logits}[L] = \dhh[L] \cdot \WU^\top$.
\item Read the top-$K$ most positive tokens (associated with input $c$) and the
   top-$K$ most negative tokens (associated with input $k$). The two poles
   together describe the contrast in token space.
\end{enumerate}

No parameters are fitted. The choices are the input pair, the read position, and $K$.
The pair should be aligned: the same token length, read at the same index, so
position-dependent structure (strong under rotary embeddings) cancels in the
difference rather than leaking into the readout. Section~\ref{sec:design} shows
this matters most at the first token, which carries a massive activation. Where
names of different lengths make exact alignment impossible
(\S\ref{sec:retrieval}), we do not simply trust the read; we verify it against a
length-matched control.

\paragraph{Two practical notes.} We project through the logit lens ($\WU$),
not the input embedding. The residual stream aligns with output space, and
projecting the same states through $W_E$ returns noise. We also project the raw
states, bypassing the final LayerNorm ($\text{LN}_f$). For a difference of two
matched states the learned shift cancels. The per-dimension rescaling preserves
the top-5 tokens (mean cosine $\approx 0.95$ against the normalized projection)
and reshuffles only the ranking tail, which does not affect an axis-level
reading. Folding the $\text{LN}_f$ gain into $\WU$, as TransformerLens does, is
equivalent.

\subsection{Per-position reading}

By reading the contrast at every position, we trace information flow: at
which position does a meaning distinction first appear? Does it appear at
the differing token itself, or at a later position that attends to it?

\subsection{Attention and MLP decomposition}

Each transformer layer adds two components to the residual stream: attention
output and MLP output. We capture both via forward hooks and project each
separately through $\WU$. This reads which of the two writes a given content
distinction at each layer, taken directly from the projection with no
trained probe.

Projecting MLP writes and attention-head writes through $\WU$ is old news: the
logit lens has been applied per-component before us. What the contrast adds is
that differencing two matched runs often exposes structure inside a component
that the raw per-component read does not show. Where a single run's MLP or head
write projects to token soup, the difference of two runs at the same component
can resolve into a legible distinction, because the shared bulk cancels and the
part that separates the two inputs is left.

\subsection{What the projection reads}

The contrastive projection reads the difference of the two states through
$\WU$. A raw logit-lens readout of either input is dominated by signals
the two inputs share (often high-frequency function words at the top of
the ranking). The subtraction cancels that shared component and reads what
remains.

$\WU$ projection assigns token labels to directions in the residual stream.
Whether those labels are interpretable varies with the contrast and the layer.
For a well-chosen contrast at a legible layer they name what differs between the
two inputs. Many other projections return token soup (\S\ref{sec:triangulation}).
The readout is a projection onto $\WU$, not a decomposition of the computation.
A component of the difference that aligns with no token direction does not
appear, and its absence from the readout is not evidence of its absence from the
state \citep{tuomi2026visibility}. Our claim is limited: for well-chosen
contrasts, part of what differs projects to legible tokens.

\subsection{Contrast design}
\label{sec:design}

What the readout surfaces depends on how the pair is built. Three rules make a
contrast legible: a shared preamble, a matched current token, and a matched
predicted next token.

\paragraph{Preambles.} \emph{Preamble}
is our name for the shared text before the contrasted change. Early tokens
are a poor place to put a contrast, for two reasons. The sharp one is
measurable: the first token carries a massive activation, residual norm about
twenty times the rest (941 versus $\sim$45 at L8), the attention-sink structure
of \citet{sun2026}. A difference on that token is amplified and dominates the
read, which is why the capitalisation mismatch below is so destructive. The 
subtler one does not show up as norm. In its first several tokens the model 
is still settling into the register and format of the text. Whatever the cause, 
a difference placed this early reads less cleanly even when its norm is ordinary: 
our bare prompts align with the preambled read at only cosine 0.85 
(Table~\ref{tab:preamble}), and shorter prompts read worse. So the rule is to place the
contrasted change several tokens in, where the read is clearest. A shared
lead-in of a few plain tokens does this.

A first-token mismatch shows why the frame matters (Table~\ref{tab:preamble}).
Read at the final token, the matched pair ``The hot dog was'' / ``The cold dog
was'' reads food at L20 (top token ``tast,'' food at rank 3), aligning with the
preambled read at cosine $0.85$. Lowercasing only the second prompt's first
token, ``The hot dog was'' / ``the cold dog was,'' pushes food down to rank 12
and puts non-food tokens on top (``sold, more''), dropping the alignment to
cosine $0.64$. A shared preamble reads food most stably (rank 0). The same
edibility contrast is read along different directions depending on the frame.

\begin{table}[htbp]
\centering
\small
\begin{tabular}{@{}ll l r r@{}}
\toprule
First prompt & Second prompt & Readout at L20 & food & cos \\
\midrule
The hot dog \underline{was}          & The cold dog \underline{was}          & tast, circular, price     & 3  & 0.85 \\
The hot dog \underline{was}          & the cold dog \underline{was}          & sold, more, circular      & 12 & 0.64 \\
$\ldots$ the hot dog \underline{was} & $\ldots$ the cold dog \underline{was} & flavor, tasted, addictive & 0  & 1.00 \\
\bottomrule
\end{tabular}
\caption{The same edibility contrast (hot vs cold), read at the final token
(underlined) under three frames: both prompts capitalised (matched), the second prompt's
first token lowercased (a mismatch), and both behind a shared preamble
($\ldots$ = ``My grandmother said that''). \emph{Readout} is the verbatim top-3
at L20; \emph{food} is the best food-token rank there; \emph{cos} is the cosine
of the L20 difference direction against the preambled one. The mismatch pushes
food down the ranking and rotates the read ($0.85\!\to\!0.64$); the preamble
reads food most stably. The mismatch degrades only the mid-stack read: by L24
all three recover to clean food.}
\label{tab:preamble}
\end{table}

Why does the mismatch hurt? The capitalization difference on its own is large
and illegible. The pure ``The'' versus ``the'' contrast (``The hot dog was''
versus ``the hot dog was,'' same words otherwise) has a difference vector of
norm about 40 at L20, and reads as junk (``NEY, ENE''). That junk is a large
fraction of the food contrast itself (norm about 68 at L20), so a start mismatch
rotates the reading toward it. The cause is position, not capitalisation: any
difference on the first token would do the same. The rule follows. Match the
prompts everywhere except the target, and keep the target several tokens past
the start with a shared lead-in. For a strong lexical contrast the preamble is only a
refinement: the food signal still recovers by the late layers, and only the
mid-stack read is muddied. 

\paragraph{Match the current token.} The token at the read position is processed
heavily by the early layers. An incidental difference there, even a nearly
meaningless one, adds a large surface component that swamps the difference of
interest. A $2\times2$ makes this precise
(Table~\ref{tab:currenttoken-prompts}): cross a context (reading a \emph{novel}
versus watching a \emph{lecture}) with a synonym at the final word (\emph{boring}
versus \emph{dull}), and read the same novel-versus-lecture difference. When the
final word matches, both poles read the context cleanly from the middle layers
(Table~\ref{tab:currenttoken-matched}): the novel pole reads ``novels, literary,
readers,'' the lecture pole reads ``lectures, seminar, students, videos.'' When
the final word is a synonym mismatch, the difference also carries the
boring-versus-dull swap. Its norm is about ten times larger at L1 (30 versus
3--5), and the swap's suffix morphology (``ly, ed, ers, ing'') dominates the
readout (Table~\ref{tab:currenttoken-mismatch}). The context is buried, and
surfaces only at the final layers (L30--32, ``readers'' on the novel pole,
``video, youtube'' on the lecture pole), even there mixed with fragments. 
Matching the current token clears the readout. This is why the trajectories 
here are read at shared-token positions
such as ``dog'' and ``was.'' The same rule applies to baseline subtraction:
snippets chosen to end in the target's final token cancel the current-token
component.

\begin{table}[htbp]
\centering
\small
\begin{tabularx}{\columnwidth}{@{}lY@{}}
\toprule
Context & Prompt \\
\midrule
novel   & I spent the whole afternoon reading the novel and found it extremely \underline{[boring / dull]} \\
lecture & I spent the whole afternoon watching the lecture and found it extremely \underline{[boring / dull]} \\
\bottomrule
\end{tabularx}
\caption{Prompts for the current-token $2\times2$. The read contrasts the two
contexts (novel versus lecture); the final word (boring versus dull) is crossed
with them, matched or mismatched, and read at the final token (underlined).}
\label{tab:currenttoken-prompts}
\end{table}

\begin{table}[htbp]
\centering
\small
\begin{tabular}{@{}lll@{}}
\toprule
L & $+$ reading the novel & $-$ watching the lecture \\
\midrule
4  & novels, literary, rette   & participants, seminar, attendees \\
20 & literary, satir, novels   & lectures, presenter, students \\
28 & readers, Readers, reader  & lectures, lecture, videos \\
32 & readers, reader, literary & lectures, lecture, teaching \\
\bottomrule
\end{tabular}
\caption{\emph{Matched} current token (both prompts end in ``dull''). The
novel-versus-lecture difference reads both poles cleanly: the novel/reading pole
($+$) and the lecture/watching pole ($-$). Verbatim top-3 (``rette'' is a
fragment).}
\label{tab:currenttoken-matched}
\end{table}

\begin{table}[htbp]
\centering
\small
\begin{tabular}{@{}rrll@{}}
\toprule
L & $\lVert d\rVert$ & $+$ reading the novel & $-$ watching the lecture \\
\midrule
4  & 34.5 & ed, ers, iveness  & ly, cium, imi \\
20 & 47.5 & ers, er, ively    & iosis, clip, cone \\
28 & 64.5 & ers, er, ership   & but, I, video \\
30 & 76.2 & ers, er, readers  & video, clip, Video \\
32 & 26.9 & er, ers, readers  & Videos, youtube, video \\
\bottomrule
\end{tabular}
\caption{\emph{Mismatched} current token (novel/dull minus lecture/boring). The
boring-versus-dull suffix morphology (ly, ed, ers, ing) dominates, and the
difference norm is many times the matched pair's. The context surfaces only at
L30--32 (``readers'' on the novel pole; ``video, youtube'' on the lecture pole),
still mixed with fragments. Verbatim top-3.}
\label{tab:currenttoken-mismatch}
\end{table}

\paragraph{Match the predicted next token.} The two rules above concern the
read position's input. A third concerns its output. The logit lens at a position
reads mostly that position's own next-token prediction. This can bury content
the position holds for later use, since downstream positions read a residual
through attention. To surface that content, choose a contrast whose two prompts
predict the \emph{same} next token at the read position. The shared prediction
then cancels, and the difference keeps only what differs downstream. Read at
``Japan'' in ``The capital of Japan'' versus ``The currency of Japan''
(Table~\ref{tab:nexttoken-prompts}), both prompts predict ``is.'' The difference splits by relation at the late layers
(Table~\ref{tab:nexttoken}): capital$-$currency reads place vocabulary,
currency$-$capital reads monetary vocabulary. France behaves the same on the
currency side (``denomination, exchange, devalue, pegged''), with a noisier
capital side (``city, location''). Now read the same contrast one token later,
at the ``is'' of ``The capital of Japan is'' versus ``The currency of Japan
is.'' Here the two prompts predict different next tokens, and the difference
reads those predictions instead: capital$-$currency reads capital cities
(``Tokyo, London, Paris, Beijing''), currency$-$capital reads currency names
(``yen, dollar, yuan''). Matching the next token reads how a position applies a
relation. Reading where the next tokens differ reads the answer that relation
produces.

\begin{table}[htbp]
\centering
\small
\begin{tabular}{@{}ll@{}}
\toprule
Capital-of & Currency-of \\
\midrule
The capital of \underline{Japan}     & The currency of \underline{Japan} \\
The capital of Japan \underline{is}  & The currency of Japan \underline{is} \\
\bottomrule
\end{tabular}
\caption{Prompts for the next-token contrast, read positions underlined. The
first pair is read at the ``Japan'' position (shared next token ``is''); the
second at the final ``is'' (differing next tokens).}
\label{tab:nexttoken-prompts}
\end{table}

\begin{table}[htbp]
\centering
\small
\begin{tabular}{@{}rlll@{}}
\toprule
L & Raw (both) & capital $-$ currency & currency $-$ capital \\
\midrule
24 & is, ,, 's  & loc, Location, locate    & currencies, denomination, exchanges \\
28 & is, ,, and & lat, obe, location       & circulated, denomination, predec \\
32 & is, ,, was & capitals, city, Population & pegged, denomination, devalue \\
\bottomrule
\end{tabular}
\caption{Matching the next token isolates a stored relation. Read at the
``Japan'' position of ``The capital of Japan'' versus ``The currency of Japan''
(Phi-2). Both prompts predict ``is,'' so the raw read is the same for each and
cancels in the difference; the difference then splits by relation, place
vocabulary on the capital side and monetary vocabulary on the currency side.
Tokens are verbatim top-3 (fragments such as ``loc,'' ``lat'' are location
sub-words; ``predec,'' ``obe'' are noise).}
\label{tab:nexttoken}
\end{table}

\subsection{Baseline subtraction: the single-prompt variant}
\label{sec:baseline}

A contrastive projection needs a second prompt. One generic alternative uses
none. From the target's residual, subtract the mean residual of many text
snippets of the same length, read at the same position. The same length matters
for a positional reason: it keeps the read at the same index in every snippet,
so position-dependent structure (strong under rotary embeddings) is shared and
cancels in the average, rather than leaking into the readout. The subtraction then removes
what is common to text of that kind and leaves what is specific to the target.
The baseline should match the target's style. A mean over English prose sentences
cancels the function-word and sentence-frame content of a prose target cleanly.
A mean over arbitrary web text (code, lists, markup) leaves a noisier remainder.
The closer the baseline's style to the target, the cleaner the readout.
Multi-contrast triangulation (\S\ref{sec:triangulation}) is the limit of this: a
baseline built from the target's own frame.

Baseline subtraction is coarser than a matched pair. A matched pair cancels the
shared computation almost exactly. The prose mean cancels only the generic
style, so the target's content clears the readout only at the late layers, where
its prediction is strong. Reading the average itself shows what is removed
(Table~\ref{tab:decomposition}). For ``My grandmother said that the hot dog
was'' (Table~\ref{tab:decomposition-prompts}), the mean of 100 English prose
sentences reads generic sentence-final
content (articles and punctuation). The difference surfaces the target's food
vocabulary.

\begin{table}[htbp]
\centering
\small
\begin{tabularx}{\columnwidth}{@{}lY@{}}
\toprule
Role & Text \\
\midrule
Target   & My grandmother said that the hot dog \underline{was} \\
Baseline & mean residual of 100 English prose sentences of the same length \\
\bottomrule
\end{tabularx}
\caption{Prompts for baseline subtraction: one target, and the prose mean it is
read against. Both read at the final token (underlined).}
\label{tab:decomposition-prompts}
\end{table}

\begin{table}[htbp]
\centering
\small
\begin{tabular}{@{}rlll@{}}
\toprule
L & Raw (target) & Prose average & Baseline-subtracted \\
\midrule
20 & not, more, made, a  & in, first, the   & not, made, always, too \\
24 & not, more, a, better & the, in, \{,\}  & delicious, better, easier, more \\
28 & more, better, too, not & the, a, \{,\} & delicious, cooked, hotter, tast \\
\bottomrule
\end{tabular}
\caption{Baseline subtraction on ``My grandmother said that the hot dog was''
(Phi-2, final position), against the mean of 100 English prose sentences of
the same length. The prose average reads generic sentence-final content
(articles, punctuation); the difference surfaces the target's food
vocabulary, which clears the readout at the late layers (L28: ``delicious,
cooked, hotter, tast'').}
\label{tab:decomposition}
\end{table}

\subsection{Multi-contrast triangulation}
\label{sec:triangulation}

A single-pair difference spans every axis on which the two prompts differ, not
only the intended one. Multi-contrast triangulation isolates the intended axis:
contrast the target against several baselines that share that axis but differ on
the incidental ones, then average the difference vectors. The incidental axes
vary across baselines and cancel; the axis common to every contrast survives. To
isolate edibility in ``hot dog,'' for instance, read it against angry, small,
and pet dog, all non-edible: edibility is the one difference every contrast
shares, so anger, size, and petness average away. It is the middle ground
between a single pair and subtracting a generic corpus average
(\S\ref{sec:baseline}): matched baselines cancel the incidental axes without
also cancelling the sentence frame.

A companion paper explains why averaging surfaces the shared content
\citep{tuomi2026visibility}. A token-aligned component surfaces in the top-$K$
only above a visibility threshold $f^{*} = 2\ln(2V)/(d + 2\ln(2V)) \approx
2\ln(2V)/d$ (the approximation holds because $d \gg \ln V$). Averaging $N$
matched difference vectors suppresses the incoherent remainder by $\sqrt{N}$,
which lowers that bar. Averaging does not sharpen the signal. It removes the
unaligned remainder. Entity-identity contrasts (a name, a city) differ on little
besides the target and read cleanly from a single pair; axes that co-vary with
several others benefit most. The result is read at the granularity the contrast
design provides, so a surfaced token may name a bundle of co-activated features
rather than a single one. We exercise triangulation where it earns its keep: in
\S\ref{sec:crossmodel} it recovers the noun-internal food read in models where a
single pair leaves it below the readout's detection bar.

\subsection{Using a contrast: decompose and test}
\label{sec:decompose}

A two-item contrast rarely lands on a single axis. When a readout looks mixed, we
take it apart: name the concept axes the pair might span, build each as its own
data-derived contrast (a mean over one noun set minus a mean over another at the
read position), and read each through $\WU$ to confirm it names its concept.
Projecting the original difference onto the set shows which axes it carries. We
then test each for causality by injecting it alone, moving one state toward the
other along that axis and measuring the effect on the prediction, rather than
trusting the raw readout. This is also why triangulation helps: averaging over
baselines that share the intended axis cancels the others. Reading a category as
a region of a space of interpretable quality dimensions is the conceptual-spaces
view of concepts \citep{gardenfors2000}, in the tradition of the semantic
differential \citep{osgood1957}. We work the method through on the hot-dog
contrast in \S\ref{sec:compound}.

\section{Lexical disambiguation}

\subsection{Compound noun: hot dog}
\label{sec:compound}

\begin{table}[htbp]
\centering
\small
\begin{tabularx}{\columnwidth}{@{}llYY@{}}
\toprule
 & Prompt & Top next-token predictions & Greedy continuation \\
\midrule
Hot dog (food)    & The hot \underline{dog} \underline{was} & too (0.088), more (0.080), cooked (0.063) & too spicy for the child \\
Cold dog (animal) & The cold \underline{dog} \underline{was} & sh[ivering] (0.447), shaking (0.036) & shivering in the snow \\
\bottomrule
\end{tabularx}
\caption{The two prompts of the compound-noun contrast, read at ``dog'' and
``was'' (underlined). The top next-token tokens are ambiguous (``too, more''); the
greedy continuation makes the food-versus-animal reading clear.}
\label{tab:compound-prompts}
\end{table}

\textbf{What the distinction decomposes into.} The contrast is meant to isolate
edibility, but its poles are different concepts: the hot pole reads food (``tast,
delicious, flavor''), the cold pole reads animal care (``grooming, Paw,
breeds''). Decomposing it (\S\ref{sec:decompose}), we build three candidate axes
at the read position, edibility, petness, and temperature
(Table~\ref{tab:axes-build}), and project the hot-cold difference onto them. It
carries all three (at L20, edibility $+21$, petness $-20$, temperature $+15$):
the compound adds edibility and suppresses the petness that ``dog'' carries, and
the prompts differ in temperature besides.

\begin{table}[htbp]
\centering
\small
\begin{tabularx}{\columnwidth}{@{}lYl@{}}
\toprule
Axis & Built as (mean $-$ mean, frame ``The X was'') & Reads (L20) \\
\midrule
Edibility (E)   & edible nouns (steak, sausage, burger, \ldots) $-$ objects (rock, chair, brick, \ldots) & delicious, cooked, tasted \\
Petness (P)     & pet animals (puppy, kitten, poodle, \ldots) $-$ the same objects & roaming, adorable, wandering \\
Temperature (T) & ``hot X'' $-$ ``cold X'' over neutral nouns (soup, coffee, water, \ldots) & steam, boiling, hotter \\
\bottomrule
\end{tabularx}
\caption{The three concept axes, each built as a multi-contrast: a mean over one
noun set minus a mean over another, at the read position. Reading each direction
through $\WU$ confirms it names its concept. Averaging over many nouns cancels
the idiosyncrasies of any single one.}
\label{tab:axes-build}
\end{table}

Which axis drives the food-versus-animal prediction? We inject each in turn,
moving the cold-dog state toward hot-dog along one axis and measuring how far the
next token moves from animal toward food (Table~\ref{tab:axes}). Edibility alone
barely helps (0.03 at L20), petness alone barely helps (0.01); neither switch
works by itself. Both together recover almost all of the effect (0.98, against
1.00 for the full difference), far more than the sum. The food/animal boundary is
diagonal in this plane: food needs high edibility and low petness at once, not
one or the other. Temperature is inert, doing nothing alone (0.00) and adding
nothing to the pair. So the contrast spans three axes but only two are causal,
and they act as a plane; this makes semantic sense, since a hot dog can be cold
or warm without affecting its hotdogness. The axes are correlated (cosine
$0.37$), so the split is not perfectly clean, they are data-derived from
particular noun sets, and this is one prompt pair in one model.

\begin{table}[htbp]
\centering
\small
\begin{tabular}{@{}r rrr rrrrr@{}}
\toprule
 & \multicolumn{3}{c}{projection} & \multicolumn{5}{c}{food fraction, injected} \\
\cmidrule(lr){2-4}\cmidrule(lr){5-9}
L & E & P & T & E & P & T & E$+$P & E$+$P$+$T \\
\midrule
12 & $+15$ & $-11$ & $+17$ & 0.00 & 0.00 & 0.00 & 0.24 & 0.21 \\
16 & $+16$ & $-14$ & $+16$ & 0.01 & 0.00 & 0.00 & 0.55 & 0.48 \\
20 & $+21$ & $-20$ & $+15$ & 0.03 & 0.01 & 0.00 & 0.98 & 0.99 \\
24 & $+43$ & $-37$ & $+18$ & 0.53 & 0.01 & 0.00 & 1.00 & 1.00 \\
\bottomrule
\end{tabular}
\caption{Decomposing ``hot dog'' $-$ ``cold dog'' into concept axes: edibility
(E), petness (P), temperature (T). \emph{Projection} is the component of the
difference on each axis. \emph{Food fraction} is $P(\text{food})/(P(\text{food})
+ P(\text{animal}))$ at the next token after moving the cold-dog run toward the
hot-dog run along that axis (E makes it more edible, P makes it less pet) at the
read position (baseline cold $0.00$, hot $1.00$; the full difference recovers
$1.00$). Neither E nor P alone crosses the
boundary, but the E$+$P plane recovers it; temperature is inert (E$+$P$+$T
equals E$+$P). Edibility alone gains potency only at the readout (L24).}
\label{tab:axes}
\end{table}

\textbf{Per-position trace.} We read at each token
(Table~\ref{tab:hotdog-trace}). ``dog'' is the same token in both prompts, so
the L0 contrast is zero. Both positions first read temperature (``hot, molten''
at L1). The food pole then separates: at ``dog'' it reads ``fried'' by L5, and
at the ``was'' prediction site it builds to ``tast, charred'' by L28. The animal
pole is the mirror image, legible only later and clearest at the readout (L28:
``whine, Paw'' at ``dog,'' ``grooming, shudder'' at ``was''). Both poles
crystallize at the prediction site as the food and animal continuations of the
two prompts. How the food signal reaches ``was,'' an MLP write at ``dog'' routed
forward by attention, is decomposed next.

\begin{table}[htbp]
\centering
\small
\begin{tabular}{@{}r ll ll@{}}
\toprule
 & \multicolumn{2}{c}{At ``dog''} & \multicolumn{2}{c}{At ``was''} \\
\cmidrule(lr){2-3}\cmidrule(lr){4-5}
L & food ($+$) & animal ($-$) & food ($+$) & animal ($-$) \\
\midrule
1  & hot, molten   & ---        & hot, fiery    & --- \\
5  & fried         & Breed      & boiled        & --- \\
20 & flavor        & ---        & tast          & euth, Paw \\
28 & vendor, stand & whine, Paw & tast, charred & grooming, shudder \\
\bottomrule
\end{tabular}
\caption{Per-position contrastive readout for hot dog versus cold dog, both
poles: food (hot dog, $+$) and animal (cold dog, $-$), cumulative read at each
position (Phi-2). Both poles start as temperature and split into food and animal
by the prediction site: the food pole is legible at ``dog'' by L5, the animal
pole only at the readout (L28). ``---'' marks a pole with no legible read yet.}
\label{tab:hotdog-trace}
\end{table}

Sub-layer decomposition at L4--L5 confirms the MLP recognizes the compound at
L4, not attention. The MLP contrastive norm at L4 (20.0) dominates the attention
norm (3.8). ``fried'' appears in the MLP output but not the attention output. L5
attention writes orthogonally to the food direction ($\cos < 0.08$ across all
contrasts tested).

\textbf{Attention routing at L5, position ``was'':}
Head 19 attends to the ``dog'' position with weight 0.911 in the hot-dog context
versus 0.735 in cold-dog. It reads the compound-noun information from ``dog'' and
writes it to ``was'' (its write is read out below).

\textbf{Multi-contrast convergence.} The food-compound direction is stable
across reference points. Five contrasts (hot dog minus cold/angry/old/pet/stray
dog) give pairwise cosine 0.72--0.84 at the dog position. The signal is about
food-compound identity, not temperature or emotion.

\textbf{The detection bar.} The visibility threshold $f^{*}$
\citep{tuomi2026visibility} makes the soup-versus-signal call quantitative: a
token clears the isotropic bath (the alignments an unaligned, random component
would produce by chance) only if its energy fraction $f = \cos^{2}(x,
u_t)$ exceeds $f^{*} = 2\ln(2V)/(d + 2\ln(2V))$, which is $0.89\%$ for Phi-2. At
the prediction site, food clears the bar where the trajectory reads cleanly
(``delicious'' at $f = 1.9\%$, $2.1\times$ the bar, by L24) and falls below it in
the middle layers, where the readout is fragmentary ($f < 0.7\%$, L10--16). At
the noun the food token is weaker: ``fried'' tops the raw logit lens at L5 but
its energy fraction is only $0.6\%$, below the bar. The logit lens ranks by an
unnormalised dot product, which a token's unembedding norm inflates independently
of alignment, so ``fried'' can head the list at the noun without the difference
vector being strongly aligned with it. It heads the printed readout, but its
alignment is below $f^{*}$, so it does not stand out from the isotropic bath: a
top rank in the logit lens is not the same as clearing the bar, and the noun read
is weak rather than robust. What clears the bar is the emission at the prediction
site, not the read at the noun. The trained probe
(\S\ref{sec:crossmodel}) confirms the distinction is present regardless, because
it does not read through this bar.

\textbf{Neuron and head detail.} The MLP write at ``dog'' is distributed. The
ten neurons pushing hardest along the food direction carry only $\sim$10\% of
it. The rest spreads across hundreds. Some write food (one writes ``cooked,
foods, dishes''), some suppress animal content (another writes ``dogs,
puppies''). The routing head L5.H19, flagged above by its attention to ``dog,''
writes ``cooked, eaten'' into ``was.'' So the food direction at the prediction
position is written by a distributed MLP population and moved forward by one
head, not by any single neuron. The read points to where this happens. Patching
(below) confirms the path is causal.

\textbf{Activation patching at multiple layers and positions} (Table~\ref{tab:hotdog-patch}):

\begin{table}[htbp]
\centering
\small
\begin{tabularx}{\columnwidth}{@{}llrY@{}}
\toprule
Patched & Layer & P(food) & Greedy continuation \\
\midrule
(baseline) & ---  & 0.16 & too spicy for the child \\
dog        & 2    & 0.00 & panting heavily, so the \\
dog        & 8    & 0.02 & more appealing to the children \\
dog        & 20   & 0.13 & too spicy for the little girl \\
hot        & 2    & 0.13 & too hot to eat, so \\
was        & 2    & 0.16 & more expensive than the hamburger \\
was        & 12   & 0.06 & placed in the oven to cook \\
was        & 20   & 0.00 & shivering in the cold weather \\
\bottomrule
\end{tabularx}
\caption{Activation patching of the hot-dog prompt: the residual at one position
and layer is replaced with the cold-dog value, then generation continues
greedily (Phi-2). P(food) is the food mass at the next token. Patching ``dog''
early destroys the food reading (``panting''), and by L20 has no effect, the
compound having moved to ``was''; patching ``was'' is the mirror, harmless early
(``the hamburger'') and destroying food late (``shivering''); patching ``hot''
never matters.}
\label{tab:hotdog-patch}
\end{table}

The patching traces the two-hop chain. (1) Patching ``dog'' early destroys the
food reading (L2: ``panting heavily''), because the compound meaning lives
there; by L20 the patch has no effect (``too spicy''), the compound having
already been routed forward. (2) Patching ``hot'' never matters (``too hot to
eat''): L0 attention has already copied it to ``dog.'' (3) Patching ``was'' is
the mirror image, harmless early (L2: ``the hamburger'') and removing food from
L12 on until it is gone late (L20: ``shivering in the cold''). This matches
attention copying the compound from ``dog'' to ``was'' in between.

\textbf{Mechanism.} Attention at L0 copies ``hot'' to ``dog.'' The MLP at L4
recognizes the compound at ``dog'' (``fried'' first appears in the MLP output).
Attention at L5 broadcasts it from ``dog'' to ``was'' (H19, attn=0.91). The
contrastive projection flagged each stage. Activation patching at three
positions and multiple layers confirms the flow. The distinction traced here is
the edibility-petness plane established at the top of this section: the trace and
patching locate where that plane is written and how it reaches the prediction
site.

\subsection{Cross-model: what generalizes and what is model-specific}
\label{sec:crossmodel}

We re-run the compound-noun decomposition on three architectures chosen to
differ where it should matter. Phi-2 has dense multi-head attention, a GELU MLP,
and a parallel residual. Pythia-1.4B (GPT-NeoX) has dense attention, GELU, and a
parallel residual. Qwen2.5-1.5B has grouped-query attention, SwiGLU, and a
sequential residual. The residual decomposition
$\text{mlp}[L] = h[L+1] - h[L] - \text{attn}[L]$ (with $\text{attn}[L]$ at the
output projection) is valid for both parallel and sequential streams, so the
attention-vs-MLP split is measured identically in all three. We read the
compound-noun prompts (Table~\ref{tab:compound-prompts}) at the noun and
prediction positions inside a shared preamble (\S\ref{sec:design}). The bare prompt is
not a fair instrument here: without the preamble Qwen2.5's prediction readout
decodes to product/commerce tokens (``priced, sold, artisan'') that resolve to
food vocabulary once the preamble stabilizes the read.

Two results follow, one robust and one architecture-specific
(Tables~\ref{tab:crossmodel} and~\ref{tab:crossmodel-reads}). \textbf{Robust.} In
all three models food content reaches the prediction position through later
attention and reads cleanly there (Table~\ref{tab:crossmodel-reads}): from L5 in
Phi-2, L14 in Pythia, and L12 in Qwen (with multilingual flavor tokens, as
expected for a multilingual model). This resolve-at-the-noun,
route-to-the-prediction shape holds across dense and GQA attention, GELU and
SwiGLU MLPs, and parallel and sequential residuals. The
greedy continuations confirm the distinction is behavioural, not an artefact of
the readout (Table~\ref{tab:crossmodel-gen}): Phi-2 and Qwen continue the hot dog
as food and the cold dog as an animal, while the smaller Pythia decodes
degenerately here.

\begin{table}[htbp]
\centering
\small
\begin{tabularx}{\columnwidth}{@{}lYY@{}}
\toprule
Model & ``The hot dog was'' \ldots & ``The cold dog was'' \ldots \\
\midrule
Phi-2 (2.7B) & too spicy for the child & shivering in the snow \\
Pythia-1.4B  & a hit \ldots            & still there, and still there \ldots \\
Qwen2.5-1.5B & a popular food item     & barking at the moon \\
\bottomrule
\end{tabularx}
\caption{Greedy continuations of the two prompts. Phi-2 and Qwen continue the
hot dog as food and the cold dog as an animal (Phi-2 matches the continuations
used elsewhere). The smaller Pythia-1.4B decodes degenerately here, repeating
itself, so its food reading shows better in the readout
(Table~\ref{tab:crossmodel-reads}).}
\label{tab:crossmodel-gen}
\end{table}

\textbf{The noun-internal read needs triangulation.} A single hot/cold pair
reads food at the noun in Phi-2, where ``fried'' enters the top ranks at L4, the
layer the MLP writes it. It does not in Pythia or Qwen, where food never reaches
the top-10. This is an instrument limit, not an absence. We average the
food-compound direction over five baselines (hot dog minus \{cold, angry, old,
pet, stray\} dog), the triangulation of \S\ref{sec:triangulation}. This recovers
a legible noun read in Pythia (best food rank 3 against 17 for the single pair,
at L15--17) and lifts Qwen from rank 60 to 20, food-service vocabulary that stops
short of the top-10 (Table~\ref{tab:crossmodel-reads}). So the compound has a token face at the noun in Phi-2 and
Pythia: read directly in one, recovered by triangulation in the other. In Qwen
it surfaces only partially. We do not claim every intermediate site carries a
clean token face, since Qwen's stays partial even after triangulation. But the
single-pair illegibility is mostly an instrument effect, not evidence that the
channel has no token form. Triangulation raises the energy fraction $f$ where a
component sits near the readout's detection bar $f^{*}$
\citep{tuomi2026visibility}: averaging the five baselines lifts the Pythia noun
read to about $1.3\times f^{*}$, while Qwen's stays below $f^{*}$ even then,
matching its partial recovery.

\textbf{A probe confirms the distinction is present, token face or not.} A
trained probe settles whether the illegible noun read is an absence or only an
unemitted computation. We train a linear probe to separate clear food nouns
(steak, burger, bacon, \ldots) from animal nouns (poodle, kitten, rabbit, \ldots)
at the noun position, then apply it to the ``dog'' of ``hot dog'' and ``cold
dog'' (Table~\ref{tab:crossmodel-probe}). In all three models it reads hot dog's
noun as food and cold dog's as animal with near-certainty ($P(\text{food})$ of
1.00, 0.99, 1.00 for hot dog and near zero for cold dog, at five-fold probe
accuracy 1.00), including Qwen, where the contrastive read surfaces no clean food
token at the noun. The compound distinction is computed in every model; the token
face is what varies. This is the trade-off with probing in miniature: the probe
is the more sensitive detector, but it needs the food-versus-animal labels,
whereas the contrastive read needs only the prompt.

\begin{table}[htbp]
\centering
\small
\begin{tabular}{@{}lccl@{}}
\toprule
Model & hot dog $P(\text{food})$ & cold dog $P(\text{food})$ & noun read, single pair \\
\midrule
Phi-2 (2.7B)  & 1.00 & 0.00 & fried (rank 0) \\
Pythia-1.4B   & 0.99 & 0.01 & --- (rank 17) \\
Qwen2.5-1.5B  & 1.00 & 0.00 & --- (rank 60) \\
\bottomrule
\end{tabular}
\caption{The compound distinction is decodable even where no token face
surfaces. A linear probe trained on clear food versus animal nouns (five-fold
accuracy 1.00 in each model) classifies the ``dog'' of ``hot dog'' as food and
of ``cold dog'' as animal in all three models. The contrastive single-pair read
at the noun is legible only in Phi-2; the computation is present regardless.}
\label{tab:crossmodel-probe}
\end{table}

\begin{table}[htbp]
\centering
\small
\begin{tabular}{lccc}
\toprule
 & Phi-2 & Pythia-1.4B & Qwen2.5-1.5B \\
\midrule
attention & dense MHA & dense MHA & GQA \\
MLP / residual & GELU / par. & GELU / par. & SwiGLU / seq. \\
\midrule
food reaches prediction site & L5 & L14 & L12 \\
\bottomrule
\end{tabular}
\caption{Architecture across three models, and the layer at which food reaches
the prediction site through later attention.}
\label{tab:crossmodel}
\end{table}

\begin{table}[htbp]
\centering
\small
\begin{tabularx}{\columnwidth}{@{}lYYY@{}}
\toprule
Model & Prediction-site read & Noun, single pair & Noun, triangulated \\
\midrule
Phi-2 (2.7B) & delicious, charred          & fried (rank 0) & fried (rank 1) \\
Pythia-1.4B  & burgers, cuisine, delicious & --- (rank 17)  & sandwiches, eaten, lunch (rank 3) \\
Qwen2.5-1.5B & flavor, flavorful, tastes   & --- (rank 60)  & toppings, grill, taste (rank 20) \\
\bottomrule
\end{tabularx}
\caption{The compound-noun food read across three models. At the prediction site
the method reads food in all three. At the noun a single pair reads food only in
Phi-2; triangulation recovers it in Pythia (rank 3) and lifts Qwen partway (rank
20). ``---'' = no legible food token in the single-pair read; best food rank in
parentheses.}
\label{tab:crossmodel-reads}
\end{table}

\subsection{Other disambiguation cases}

The same method reads disambiguation driven by a verb sense or a particle,
the two poles giving the two senses (Table~\ref{tab:disambig}).

\begin{table}[htbp]
\centering
\small
\begin{tabularx}{\columnwidth}{lYY}
\toprule
Contrast & Pole A reads & Pole B reads \\
\midrule
``He caught a \textbf{cold}'' / ``\ldots\ \textbf{fish}'' (verb sense) & fever, coughing, cough, flu & proudly, reel, bait, trout \\
``He was \textbf{fired up}'' / ``\ldots\ \textbf{fired}'' (particle) & excited, ready, energetic & sued, blacklist, lawsuits \\
\bottomrule
\end{tabularx}
\caption{Verb-sense and particle disambiguation read at the final position
(L28). Each pole reads its sense: illness versus fishing, enthusiasm versus
dismissal.}
\label{tab:disambig}
\end{table}

A noun can also be disambiguated by a modifier. Contrasting ``The steep bank
was'' against ``The closed bank was,'' read at the shared word ``bank'' (the
current-token rule of \S\ref{sec:design}, since the modifiers differ), the two poles read the
two senses of the noun (Table~\ref{tab:bank}): the steep pole reads the
riverbank sense (``slopes, erosion, steep''), the closed pole the financial
sense (``doors, branch, closed'').

\begin{table}[htbp]
\centering
\small
\begin{tabular}{@{}rll@{}}
\toprule
L & steep pole (riverbank) & closed pole (financial) \\
\midrule
20 & slopes, slope, surrounding & Frankfurt, entities \\
24 & slopes, erosion, steep     & doors, closed, Doors \\
28 & slopes, slope, erosion     & branch, window \\
\bottomrule
\end{tabular}
\caption{A noun disambiguated by a modifier: ``The steep bank was'' versus ``The
closed bank was,'' read at the shared ``bank'' (Phi-2). The steep pole reads the
riverbank sense, the closed pole the financial sense.}
\label{tab:bank}
\end{table}

\section{Metaphor: domain routing, not a figurativity flag}
\label{sec:metaphor}

Metaphor is a clean test of what the readout can see. Does the model mark
figurative language with a single ``figurativity'' feature, or with something
more local? Four adjectives (cold, sharp, bright, heavy) are each used once
literally and once metaphorically, in a shared frame that ends just before the
two senses diverge (Table~\ref{tab:metaphor-prompts}).

\begin{table}[htbp]
\centering
\small
\begin{tabularx}{\textwidth}{@{}lYY@{}}
\toprule
Word & Literal prompt & Metaphorical prompt \\
\midrule
cold   & The ice in the bucket was extremely cold. The temperature \underline{was} & The reception at the party was extremely cold. The atmosphere \underline{was} \\
sharp  & The razor was extremely sharp. The blade \underline{was} & The review was extremely sharp. The tone \underline{was} \\
bright & The spotlight was extremely bright. The light \underline{was} & The class was extremely bright. The child \underline{was} \\
heavy  & The barbell was extremely heavy. The weight \underline{was} & The room was extremely heavy. The mood \underline{was} \\
\bottomrule
\end{tabularx}
\caption{The four literal/metaphorical prompt pairs, each read at the final
token (``The [role] was,'' underlined).}
\label{tab:metaphor-prompts}
\end{table}

Each prompt is built to end at ``The [role] was,'' and
we read at that final token, a few tokens past the adjective. The predicted next token
is meant to differ: ``The temperature was'' predicts a temperature,
``The atmosphere was'' predicts a mood. Each pole then reveals the domain it routes to. At L24 the literal pole returns
the physical domain and the metaphorical pole a specific \emph{target} domain,
different for each word (Table~\ref{tab:metaphor}).

\begin{table}[htbp]
\centering
\small
\begin{tabular}{lll}
\toprule
Word & Literal pole & Metaphorical pole \\
\midrule
cold   & higher, Celsius, Fahrenheit & tense, atmosphere, mood, vibe \\
sharp  & blade, blades, stainless    & tone, sarcastic, condescending \\
bright & blinding, intensity, harsh  & proud, gifted, amazed, grades \\
heavy  & load, exert, Hercules       & mood, tense, gloomy, bleak \\
\bottomrule
\end{tabular}
\caption{Contrastive readout at L24 for the four literal/metaphorical
contrasts. Each metaphorical pole routes to its own target domain: cold and
heavy to emotion, sharp to social tone, bright to intelligence.}
\label{tab:metaphor}
\end{table}

The routing direction is stable across sentences for each word (pairwise cosine
0.64--0.88 over four pairs). To check it is not memorised from the particular
prompts, we extract it from three pairs and inject it into the held-out fourth
(leave-one-out). Table~\ref{tab:metaphor-loo} shows the four folds for cold.
Injecting toward emotion on a held-out literal prompt surfaces ``tense,
chilly.'' Injecting toward temperature on a held-out metaphorical prompt
surfaces ``below, freezing.'' This holds in every fold. The effect is graded,
and generating a few tokens after the injection shows it
(Table~\ref{tab:metaphor-gen}). At unit magnitude only cold reverses to its
literal domain (``below freezing''); doubling the injection reaches the literal
domain for sharp (``sharp'') and bright (``blinding'') as well. Heavy is the
exception, staying at ``too dark,''.
So the routing direction is domain-specific for all four (the cosine matrix
below); the magnitude needed to flip a whole continuation just varies by word.

\begin{table}[htbp]
\centering
\small
\begin{tabularx}{\columnwidth}{cYY}
\toprule
Held-out fold & Literal prompt, $-$dir (toward emotion) & Metaphorical prompt, $+$dir (toward temperature) \\
\midrule
0 & tense, chilly, freezing & so, below, freezing \\
1 & tense, icy, so          & so, below, very \\
2 & chilly, tense, unw      & so, below, much \\
3 & so, chilly, tense       & very, below, freezing \\
\bottomrule
\end{tabularx}
\caption{Leave-one-out for cold. The routing direction is built from three
literal/metaphorical pairs and injected ($\pm 1\times$) into the held-out
fourth. Top-3 readout tokens after injection; the direction shifts a
held-out literal prompt toward emotion and a held-out metaphorical prompt
toward temperature, so it is not memorised from the extraction prompts.}
\label{tab:metaphor-loo}
\end{table}

\begin{table}[htbp]
\centering
\small
\begin{tabularx}{\columnwidth}{@{}lYY@{}}
\toprule
Word & Literal direction $+1\times$ & $+2\times$ \\
\midrule
cold   & below freezing, and everyone & below freezing \\
sharp  & very negative                & sharp \\
bright & too young to understand      & blinding \\
heavy  & too dark for the party       & too dark for the party \\
\bottomrule
\end{tabularx}
\caption{Greedy continuations of each word's metaphorical prompt with the
literal routing direction injected at $1\times$ and $2\times$ (Phi-2, L24). At
$1\times$ only cold reverses to its literal domain (``below freezing''); at
$2\times$ sharp (``sharp'') and bright (``blinding'') reach it too. Heavy stays
at ``too dark,'' a mood word, not its literal weight domain.}
\label{tab:metaphor-gen}
\end{table}

Injection is bidirectional and graded. Adding the cold direction to ``The
reception was extremely cold. The atmosphere was'' shifts ``chilly'' (social)
toward ``below'' (temperature), with the top-1 flip at $0.65\times$ its natural
magnitude. Subtracting it from the literal prompt shifts ``below'' toward
``tense'' (emotion) by $1.5\times$.

The decisive test is cross-domain. If the model had one literal/figurative axis,
a word's routing direction should move any context toward that context's own
literal sense. It does not (Table~\ref{tab:metaphor-cross}). Injected into a
sharp tone or a heavy mood, the cold direction drives the continuation to
temperature (``below zero''), not toward sharpness or weight. The sharp direction
does the reverse, driving a cold atmosphere or a heavy mood to ``sharp.'' Each
direction imposes its own target domain whatever the context, so the mapping is
source-specific, not a generic figurativity flag.

\begin{table}[htbp]
\centering
\small
\begin{tabularx}{\columnwidth}{@{}lYY@{}}
\toprule
Direction $\rightarrow$ context & Baseline & Injected ($3\times$) \\
\midrule
cold $\rightarrow$ ``The tone was'' \emph{(sharp)}       & very negative        & below zero \\
cold $\rightarrow$ ``The mood was'' \emph{(heavy)}       & somber               & below zero \\
sharp $\rightarrow$ ``The atmosphere was'' \emph{(cold)} & chilly, unwelcoming  & sharp and unwelcoming \\
sharp $\rightarrow$ ``The mood was'' \emph{(heavy)}      & somber               & sharp \\
\bottomrule
\end{tabularx}
\caption{Cross-domain injection at $3\times$ (Phi-2, L24): one word's routing
direction injected into another word's context, greedy continuation. Each
direction imposes its own target domain regardless of the context. The cold
direction drives a sharp tone or a heavy mood to temperature (``below zero''),
the sharp direction drives a cold atmosphere or a heavy mood to sharpness
(``sharp''). A single literal/figurative axis would instead move each context
toward its own literal sense.}
\label{tab:metaphor-cross}
\end{table}

The cross-domain cosines agree (Table~\ref{tab:metaphor-cos}). Cold and
heavy both map a physical scale onto emotion and share structure (0.58). Bright
maps onto intelligence and is nearly orthogonal to the rest (0.03--0.14).

\begin{table}[htbp]
\centering
\small
\begin{tabular}{lrrrr}
\toprule
       & cold & sharp & bright & heavy \\
\midrule
cold   & 1.00 & 0.28 & 0.06 & 0.58 \\
sharp  & 0.28 & 1.00 & 0.03 & 0.37 \\
bright & 0.06 & 0.03 & 1.00 & 0.14 \\
heavy  & 0.58 & 0.37 & 0.14 & 1.00 \\
\bottomrule
\end{tabular}
\caption{Pairwise cosine between the four routing directions at L24.
Directions that share a target domain (cold, heavy) correlate; bright, which
routes to a different target, is orthogonal to all.}
\label{tab:metaphor-cos}
\end{table}

We read this as a claim about representation, scoped to these four words. We
find no single figurativity feature. What we call metaphor here is a set of
domain-to-domain mappings, and the projection reads one mapping at a time. The
direction that separates a literal from a figurative use is fixed by the source
and target domains it connects, not by figurativity itself. This is why a single
literal/metaphorical axis is inconsistent across words, and why the reading is
legible only when the contrast holds the domain pair fixed.

\section{Recall versus hallucination}
\label{sec:retrieval}

When the model hallucinates, it produces a confident but fabricated answer for
a fictional entity. The contrastive projection shows what its retrieval surfaces
in token space: specific facts for real entities, and nothing beyond name
fragments for fictional ones.

\textbf{Design:} We construct matched pairs where one prompt elicits genuine recall
and the other elicits hallucination, keeping the frame identical (Table~\ref{tab:halluc-prompts}):

\begin{table}[htbp]
\centering
\small
\begin{tabularx}{\textwidth}{@{}lYY@{}}
\toprule
 & Prompt & Prediction \\
\midrule
Real      & Nikola Tesla, born in 1856, invented \underline{the} & Tesla coil\ldots \\
Fictional & Ludvig von Vogelkirche, born in 1859, invented \underline{the} & first practical electric motor\ldots \\
\midrule
Real      & Marie Curie, born in 1867, \underline{discovered} & the elements polonium and radium\ldots \\
Fictional & Helena Brandström, born in 1871, \underline{discovered} & a new species of moth\ldots \\
\bottomrule
\end{tabularx}
\caption{Matched real and fictional entity prompts, read at the final token
(underlined). Both elicit confident, specific predictions.}
\label{tab:halluc-prompts}
\end{table}

Both sides produce confident, specific answers. But the contrastive projection
at L28 reads different content on each pole (Table~\ref{tab:halluc-poles}):

\begin{table}[htbp]
\centering
\small
\begin{tabular}{lll}
\toprule
Pair & Real pole (L28) & Fictional pole (L28) \\
\midrule
Tesla vs Vogelkirche & Tesla, alternating, Altern, electric & Vog, v, von, Von \\
Curie vs Brandström & radio, Radio, Radiation, radioactive & Brand, M, H, a \\
\bottomrule
\end{tabular}
\caption{Contrastive projection at L28: the real pole reads factual associations, the fictional pole only name fragments.}
\label{tab:halluc-poles}
\end{table}

The real pole reads \emph{factual associations} (Tesla $\rightarrow$ alternating
current, Curie $\rightarrow$ radioactivity). The fictional pole reads \emph{name
fragments} (Vog, von, Brand). The readout contains nothing about the entity
beyond the tokens of its name.

\textbf{Confirmation via entity-versus-generic contrast.} Contrasting each
entity against a bare frame (``A person, born in [year], invented the'' /
``\ldots\ discovered'') isolates what the name adds (Table~\ref{tab:halluc-generic}).
The real names add factual content (Tesla: alternating current; Curie:
radioactivity); the fictional names add only fragments of themselves. The
hallucinated entity carries no factual content beyond its name.

\begin{table}[htbp]
\centering
\small
\begin{tabularx}{\columnwidth}{@{}lY@{}}
\toprule
Entity $-$ generic frame & Read at L28 \\
\midrule
Nikola Tesla \emph{(real)}          & alternating, Tesla \\
Ludvig von Vogelkirche \emph{(fictional)} & Vog, von, Von \\
\midrule
Marie Curie \emph{(real)}           & radio, Radio, rad \\
Helena Brandström \emph{(fictional)}   & Brand, brand \\
\bottomrule
\end{tabularx}
\caption{Each entity minus a generic frame at L28, isolating what the name adds.
Real names add factual content (Tesla, alternating current; Curie,
radioactivity); fictional names add only fragments of themselves.}
\label{tab:halluc-generic}
\end{table}

\textbf{Contrastive norm.} The relative norm $\|\dhh\|/\|h\|$ at L28 is larger
for real-versus-fictional pairs (mean 0.98) than real-versus-real pairs (mean
0.70). Two real entities differ less from each other than either differs from a
fictional one.

\textbf{Length-matched control.} These entity pairs are not token-length matched:
``Nikola Tesla'' and ``Ludvig von Vogelkirche'' differ by seven tokens, so the
read token sits at a different index and rotary position structure need not
cancel. The first-token mismatch of \S\ref{sec:design} is the sharp version of
this failure; a mid-prompt length offset, read at the final token, is milder, but
it still warrants a check. We rebuild the contrast with real and invented names
of identical token count in the same frame (Table~\ref{tab:halluc-matched}). The
finding holds: the real pole reads the entity's factual associations (Tesla,
alternating current; Curie, radioactivity; Newton, gravity), the fictional pole
only name fragments and junk. The effect is not an artifact of the length
mismatch. The contrastive-norm gap above, being a magnitude, is the claim most
exposed to it, since real-versus-fictional pairs also carry the larger length
differences; we read that number as suggestive rather than exact.

\begin{table}[htbp]
\centering
\small
\begin{tabularx}{\columnwidth}{@{}lYY@{}}
\toprule
Real vs invented (matched length) & Real pole (L28) & Fictional pole (L28) \\
\midrule
Tesla vs Kessler (11 tok)  & Tesla, Altern, alternating & Kessler, tiss, helicop \\
Curie vs Kessler (10 tok)  & Radio, Radiation, rad      & Kessler, lake, H \\
Newton vs Kessler (10 tok) & Newton, gravity, Laws      & mineral, Targ, Koen \\
\bottomrule
\end{tabularx}
\caption{Length-matched control for Table~\ref{tab:halluc-poles}. Each real
entity is paired with an invented name (Halvard Kessler) of identical token
count, same frame, read at the shared final token (Phi-2, L28). The real pole
reads factual associations, the fictional pole only name fragments and junk, as
in the unmatched pairs; the finding does not depend on the length mismatch.}
\label{tab:halluc-matched}
\end{table}

\textbf{Entropy.} The real inventors predict with lower entropy (H = 1.5--2.5)
than their fictional counterparts (H = 3.4--6.3), consistent with the model
having specific knowledge to draw on. But entropy alone cannot separate
confident recall from confident hallucination. The mountain case shows this
(Tables~\ref{tab:mountain-prompts} and~\ref{tab:mountain}). The real Mount Cook
and the fictional Mount Silverhorn both predict a confident, specific height at almost the same entropy
(2.14 versus 2.36), so entropy cannot tell them apart. The projection can: each
real mountain minus Silverhorn reads geographic knowledge (Cook ``volcano,
climbers, Peaks,'' Everest ``Nepal, Tibet, Himalaya''), while Silverhorn reads
only generic number-range priors (``1100, 1200, 1300'').

\begin{table}[htbp]
\centering
\small
\begin{tabularx}{\columnwidth}{@{}lY@{}}
\toprule
Mountain & Prompt \\
\midrule
Everest \emph{(real)}          & Mount Everest, the tallest peak in the Himalayas, rises \underline{to} \\
Cook \emph{(real)}             & Mount Cook, the tallest peak in New Zealand's Southern Alps, rises \underline{to} \\
Silverhorn \emph{(fictional)}  & Mount Silverhorn, the tallest peak in New Zealand's Southern Alps, rises \underline{to} \\
\bottomrule
\end{tabularx}
\caption{Prompts for the mountain case; read at the final token (underlined).}
\label{tab:mountain-prompts}
\end{table}

\begin{table}[htbp]
\centering
\small
\begin{tabular}{@{}llll@{}}
\toprule
Mountain & Entropy & Predicts & Contrastive pole \\
\midrule
Mount Everest \emph{(real)}       & 1.23 & 29{,}032\,ft        & Nepal, Tibet, Himalaya \\
Mount Cook \emph{(real)}          & 2.14 & 3{,}724\,m          & volcano, climbers, Peaks \\
Mount Silverhorn \emph{(fictional)} & 2.36 & $\approx$2{,}800\,m & 1100, 1200, 1300 \\
\bottomrule
\end{tabular}
\caption{The mountain case (Phi-2, ``\ldots\ rises to''). \emph{Entropy} is at
the next (height) token; \emph{Predicts} is the greedy height. Cook (real) and
Silverhorn (fictional) both emit a confident height at nearly the same entropy,
so entropy cannot tell them apart: Cook recalls its real 3{,}724\,m, Silverhorn
invents one. \emph{Contrastive pole} is that mountain's side of the real-minus-fictional
contrast (real poles at L24, Silverhorn at L28): the real mountains read
geographic knowledge, the fictional one reads only generic number-range priors.}
\label{tab:mountain}
\end{table}

The contrastive projection does not find a ``hallucination flag.'' It reads what
retrieval surfaces in token space, and for fictional entities that is only name
tokens and contextual priors. One caveat applies to every absence reading: a
token missing from the top-$K$ has an energy fraction $f$ below the readout's
visibility threshold $f^{*}$, which bounds the signal but is not evidence of zero
signal \citep{tuomi2026visibility}. What separates the fictional entities is that no
factual content surfaces at any layer, while for the matched real entities it
does.

\section{What the readout reads}
\label{sec:readout}

Every reading so far has named a distinction in specific tokens: ``fried,
crispy'' for food, ``Nepal, Tibet'' for a real mountain. Those tokens are the
readout's output, not the computation. So one question decides how to use them.
Is a particular token identity a fact about what the model computes, or about
the basis this one network happened to learn? A seed control answers it. We read
four contrasts (Table~\ref{tab:seed-prompts}) through five Pythia-410M models
that share architecture, tokenizer, and training data and differ only in random
initialization seed: the base model and four of the seed reruns released in the
Pythia suite (seeds 1, 3, 6, and 7). The particular seeds are incidental to the
finding, which needs only that the networks differ solely in initialization.

\begin{table}[htbp]
\centering
\small
\begin{tabularx}{\textwidth}{@{}lYY@{}}
\toprule
Contrast & Positive prompt & Negative prompt \\
\midrule
Sentiment    & The film was absolutely wonderful and \underline{I}      & The film was absolutely terrible and \underline{I} \\
Temperature  & I touched the metal and it felt extremely \underline{hot} & I touched the metal and it felt extremely \underline{cold} \\
Time         & This will definitely happen \underline{tomorrow}          & This definitely happened \underline{yesterday} \\
Nationality  & She grew up in Paris speaking fluent \underline{French}    & She grew up in Tokyo speaking fluent \underline{Japanese} \\
\bottomrule
\end{tabularx}
\caption{The four contrasts read through the five seeds, each at the final
token (underlined). Table~\ref{tab:seedface} shows the per-seed reads.}
\label{tab:seed-prompts}
\end{table}

\paragraph{The token-face is network-specific.} At the read layer the top-10
tokens barely overlap across seeds (mean pairwise Jaccard $0.08$; about 45
distinct tokens fill the 50 top-ranked slots). Yet the contrastive direction
boosts the concept anchors over their opposites in every seed ($20/20$
seed$\times$contrast, margins $+2.8$ to $+5.9$). Table~\ref{tab:seedface} shows
all four contrasts across the five seeds: the top tokens differ from seed to
seed, and on this small model are often fragments, yet the concept margin stays
positive throughout. Nationality reads Francophone geography (different members
per seed) and time reads future words; sentiment and temperature surface mostly
seed-specific fragments. The same holds across architectures. The compound-noun food
reading is ``delicious'' in Phi-2, ``burgers, cuisine'' in Pythia, and ``flavor,
flavorful'' in Qwen (\S\ref{sec:crossmodel}). Same field, different tokens.

\begin{table}[htbp]
\centering
\small
\begin{tabularx}{\columnwidth}{@{}lYYYY@{}}
\toprule
Seed & Sentiment & Temperature & Time & Nationality \\
\midrule
base  & aug, agus            & ter, ps   & someday, unless   & Alger, France \\
seed1 & capturing, relaxation & ter, ting & someday, morrow  & Alger, Qu\'{e}bec \\
seed3 & capture, captures    & ta, te    & tom, whichever    & Alger, Belgium \\
seed6 & tail, tails          & ter, pone & morrow, someday   & Alger, France \\
seed7 & empre, adors         & ter, tera & tom, dat          & Alger, France \\
\bottomrule
\end{tabularx}
\caption{The four contrasts read through five Pythia-410M models differing only
in initialization seed (top-2 legible tokens, positive pole, at each seed's read
layer). The tokens differ from seed to seed and, on this small model, are often
sub-word fragments (mean pairwise top-10 Jaccard $0.005$--$0.11$). Yet the
contrastive direction boosts the concept anchors over their opposites in every
seed and contrast ($20/20$; margins $+2.8$ to $+5.9$): the distinction is shared
even where the token-face is not. Nationality reads Francophone geography and
time reads future words most legibly; sentiment and temperature surface mostly
seed-specific fragments.}
\label{tab:seedface}
\end{table}

\paragraph{What this implies.} A component of a write is token-shaped where it
aligns with $\WU$, and that component is real: the same distinction is
recoverable in every seed. But the token basis a network lands on is a property
of that trained network, not a universal code. What a computation looks like in
token space is network-specific; the distinction it draws is not. The practical
rule is to read the axis, not the tokens: confirm an axis by an anchor margin or
an injection, not by the identity of the top-ranked tokens. This network-specific
basis is one of three reasons a token can be absent from a given readout; the
other two, a below-threshold magnitude and a distinction the model never emits,
are limits of the readout itself, and the Discussion takes them up. A full
account of how token-shaped writes superpose in the residual is beyond our scope.

\section{Discussion}

The contrastive projection reads through $\WU$, so it sees only what the model
emits into the vocabulary. Section~\ref{sec:readout} gave one reason a
distinction can be missing from a readout, a network-specific basis, where the
distinction is emitted but wears different tokens in each network. Two further
reasons are limits of the readout itself. The first is magnitude: the aligned
component sits below the visibility threshold, and averaging more baselines can
raise it \citep{tuomi2026visibility}, as triangulation did for the Pythia noun
read. The second is harder. A distinction the model computes but never emits into
the output basis would be invisible to any $\WU$ readout, however clean the
contrast. Whether such computed-but-unemitted distinctions are common, and how to
detect them without the readout, is an open question. It marks the ceiling of the
method: contrastive projection reads emitted structure, not the full internal
state.

\subsection{A hypothesis: metaphor as a linear conceptual mapping}

The metaphor readings (\S\ref{sec:metaphor}) suggest a hypothesis about the
representation, beyond what four words in one model can establish. The model
appears to have no single ``figurativity'' feature. It may instead represent a
metaphorical use as the literal representation displaced by an approximately
linear offset toward the metaphor's target domain, $h(\text{metaphorical})
\approx h(\text{literal}) + v(\text{source}\rightarrow\text{target})$. Three
observations point this way. The offset is stable across sentences for each word,
a property of the mapping rather than the prompt (pairwise cosine
$0.64$--$0.88$). It is dominated by the target concept, so words that map onto
the same target share it: cold and heavy, both onto emotion, correlate at cosine
$0.58$, while bright, onto intelligence, is orthogonal. And it can be added to
induce the metaphor or subtracted to undo it, though more readily for some words
(cold at unit magnitude) than others (heavy resists even at $2\times$). This
reads as a mechanistic version of conceptual metaphor theory
\citep{lakoff1980}, with the source-to-target mapping realized as a linear
direction, in the spirit of linear relation decoding \citep{hernandez2024}.

The clean test is compositional: build a target-domain direction (emotion, say)
from cues unrelated to these words, and check whether injecting it induces the
metaphor across new source words. If the target component is genuinely shared and
separable, this should work; the graded result, where heavy resists, already
suggests the separation is imperfect. We leave this to future work.

\subsection{Future directions}

Several extensions follow directly from the method.
\begin{itemize}[leftmargin=*]
\item \textbf{A developmental clock.} Reading the same contrast across training
  checkpoints would date when a distinction becomes legible in token space,
  separating when a behaviour forms from when it enters the vocabulary basis.
  This needs a checkpointed model such as Pythia.
\item \textbf{An automatic visibility gate.} We apply the visibility threshold
  $f^{*}$ \citep{tuomi2026visibility} by hand to one example (\S\ref{sec:compound}).
  Wiring it into the readout would flag every projection automatically, replacing
  the qualitative soup-versus-signal judgement with a quantitative one.
\item \textbf{Argument-general relation directions.} The next-token contrasts
  here read a relation applied to a single argument. Triangulating a relation
  contrast over many arguments (the capital of France, of Japan, of Egypt)
  would cancel the argument and leave a direction that reads the relation
  itself. This is the natural bridge to function vectors
  \citep{todd2024}: a readout of the relation direction that neither extracts
  nor injects a vector.
\item \textbf{Reading the corpus, not the capability.} Because the $\WU$ basis
  is training-specific (\S\ref{sec:crossmodel}), the same contrast read across
  models probes what each model absorbed rather than what it can do. Qwen reads
  ``hot dog'' as a priced product where Phi-2 reads food. A systematic version
  would compare how differently-trained models frame shared concepts.
\end{itemize}

\subsection{Limitations}

\begin{itemize}[leftmargin=*]
\item \textbf{Curated pairs, not sampled.} All demonstrations use hand-constructed
  minimal pairs. The multi-contrast triangulation uses hand-selected
  baselines.
\item \textbf{LayerNorm bypassed.} We skip the final LayerNorm, so $\WU$ receives
  vectors at the wrong scale. Token rankings are empirically invariant to this.
  But raw contrastive norms are not comparable across layers, because the
  residual-stream norm grows with depth.
\item \textbf{$\WU$ readability not guaranteed.} The difference of two states was
  never trained for $\WU$ projection. Token labels at intermediate layers
  are $\WU$'s nearest-neighbour assignments. A quantitative criterion for
  when a component of the difference vector surfaces in this readout is
  developed in a companion paper \citep{tuomi2026visibility}.
\item \textbf{Smoothness is not $\WU$-specific.} Trajectory coherence
  (consecutive-layer cosine) is a property of $\dhh$, not of $\WU$.
\item \textbf{Exploratory, not causal.} The per-position trace and per-head
  decomposition identify large contributors, not causes. We treat the
  reading as a pointer and verify causality only where we make a causal
  claim: activation patching for the compound-noun circuit, and the concept-axis
  injection (\S\ref{sec:compound}). Both interventions have known failure modes we
  do not rule out: a subspace can look causal without being the mechanism
  \citep{makelov2023}, which bears on the concept-axis injection, and self-repair
  can compensate for a patched component and mask its effect \citep{mcgrath2023},
  which our single-layer patches do not control for. So we report the patching as
  a qualitative trace, not a quantitative causal measurement. The general
  causal status of the difference vector is inherited from RepE, not
  re-established here. We ran further causal tests during this work (adding and
  subtracting the difference vector to steer generation across several of the
  contrasts above), and they behaved as expected. We omit them because they only
  reproduce what RepE, ActAdd, and Contrastive Activation Addition (CAA) already establish for matched-pair
  subtraction; nothing about steering is new here, and the paper's contribution
  is the readout, not the intervention.
\item \textbf{Coherence is judged by inspection.} Every projection returns
  tokens. Whether they form a coherent reading or token soup is a
  qualitative judgement. The visibility threshold
  \citep{tuomi2026visibility} bounds when an aligned component surfaces
  in the top-$K$, but provides no automatic coherence criterion.
\item \textbf{Triangulation coverage.} We tested multi-contrast triangulation on
  6 cases. We do not know whether every model computation yields a
  token-readable component under contrastive subtraction, or how many
  baselines are sufficient in general.
\item \textbf{Per-position reading requires tokenization alignment.} The read
  position must correspond to the same structural role in both inputs.
\item \textbf{Model coverage.} Mechanistic depth is on Phi-2; the
  compound-noun circuit is re-run on Pythia-1.4B and Qwen2.5-1.5B
  (\S\ref{sec:crossmodel}), and the seed control uses five Pythia-410M
  models.
\end{itemize}

\section{Related work}

\textbf{Contrastive activation methods.} RepE~\citep{zou2023},
ActAdd~\citep{turner2023}, and CAA~\citep{rimsky2024} use matched-pair
subtraction for steering. \citet{du2026} introduced the operation we build on:
decoding an activation \emph{difference} through the logit lens, to trace
meta-cognitive control in R1-style models. We develop that primitive into a
systematic tracer (per-position, per-head, and sub-layer readout; multi-contrast
triangulation; causal checks) across a range of contrasts and three
architectures. \citet{li2024} use contrastive pairs to study truthfulness;
\citet{ma2026} contrast contextualized and non-contextualized logits to rectify
conflict-inducing layers. Both intervene; we use the contrast only to read.

\textbf{Logit and tuned lens.} The logit lens~\citep{nostalgebraist2020} and
tuned lens~\citep{belrose2023} project individual states through $\WU$; Future
Lens~\citep{pal2023} shows a single state also carries information about tokens
several positions ahead. The contrastive projection reads the content that
\emph{differs} between two inputs, a different subspace from the logit lens on
either input alone.

\textbf{Probing and linear representations.} Linear probes~\citep{belinkov2022}
train classifiers on hidden states to detect features; \citet{burns2023} extract
truth directions from contrast pairs without supervision. The contrastive
subtraction is a zero-shot linear probe along the axis the input pair defines,
read out in token space rather than through a trained classifier, and
triangulation extends it to arbitrary semantic axes.

\textbf{Superposition and sparse autoencoders.}~\citet{elhage2022} characterized
superposition in toy models; \citet{bricken2023} and~\citet{templeton2024}
decompose superposed representations into monosemantic features with sparse
autoencoders. We do not: triangulation cancels the content that varies across
baselines and reads what is left, a \emph{reading} rather than a decomposition
(the surviving component may itself be a bundle, and is read only where it
projects to $\WU$).~\citet{lange2026} note that the sparsity objectives training
cross-layer transcoders can reward circuits that rewrite deep computation into
shallow form; our reading has no learned dictionary or sparsity penalty and so
avoids that incentive, but also recovers no circuit topology and claims no
faithfulness.

\textbf{Circuit analysis and causal tracing.}~\citet{meng2022} localized factual
associations by causal tracing, \citet{gould2024} identified successor heads from
attention patterns, and \citet{conmy2023} automated circuit discovery. Our method
flags the same kind of structure (which layers, heads, and content) from the
readout alone, as for the compound-noun circuit (\S\ref{sec:compound}), but does
not establish causality; it is an exploratory complement to these techniques.

\textbf{Factual recall and relational knowledge.}~\citet{geva2023} traced factual
associations to MLP layers, where the subject's last token is enriched and the
relation applied at the final position. \citet{hernandez2024} showed many
relations are well approximated by a linear map, and \citet{todd2024} that a task
or relation is carried by a compact, causal ``function vector'' that, added to a
new context, produces the answer. The relation-versus-answer split in our
next-token note (Table~\ref{tab:nexttoken}) is the observational counterpart:
reading at the shared-next-token entity position surfaces the relation-conditioned
content, reading at the answer position the extracted attribute. We neither
extract nor inject a vector, and our per-pair difference is argument-specific, so
this reads the distinction rather than recovering a function vector. The
hallucination readings connect here too: for fictional entities nothing beyond
name tokens surfaces.

\section*{Use of AI Assistants}

Large language models (Claude, Anthropic; Gemini, Google) were used as
assistive tools for coding, running experiments, and drafting text. All
research questions, experimental design, and reported claims were
directed and verified by the author, who takes full responsibility for
the content.

\section*{Code and Data}

Code and data are available at
\url{https://github.com/EvidentSolutions/llm-interp/tree/main/contrastive}
and archived at \url{https://doi.org/10.5281/zenodo.20843136}.

\bibliographystyle{plainnat}
\bibliography{paper_v3}

\begin{thebibliography}{30}
\providecommand{\natexlab}[1]{#1}
\providecommand{\url}[1]{\texttt{#1}}
\expandafter\ifx\csname urlstyle\endcsname\relax
  \providecommand{\doi}[1]{doi: #1}\else
  \providecommand{\doi}{doi: \begingroup \urlstyle{rm}\Url}\fi

\bibitem[Belinkov(2022)]{belinkov2022}
Yonatan Belinkov.
\newblock Probing classifiers: Promises, shortcomings, and advances.
\newblock \emph{Computational Linguistics}, 48\penalty0 (1), 2022.

\bibitem[Belrose et~al.(2023)Belrose, Ostrovsky, McKinney, Furman, Smith,
  Halawi, Biderman, and Steinhardt]{belrose2023}
Nora Belrose, Igor Ostrovsky, Lev McKinney, Zach Furman, Logan Smith, Danny
  Halawi, Stella Biderman, and Jacob Steinhardt.
\newblock Eliciting latent predictions from transformers with the tuned lens.
\newblock \emph{arXiv preprint arXiv:2303.08112}, 2023.

\bibitem[Bricken et~al.(2023)]{bricken2023}
Trenton Bricken et~al.
\newblock Towards monosemanticity: Decomposing language models with dictionary
  learning.
\newblock Transformer Circuits Thread, Anthropic, 2023.

\bibitem[Burns et~al.(2023)]{burns2023}
Collin Burns et~al.
\newblock Discovering latent knowledge in language models without supervision.
\newblock In \emph{International Conference on Learning Representations
  (ICLR)}, 2023.

\bibitem[Conmy et~al.(2023)]{conmy2023}
Arthur Conmy et~al.
\newblock Towards automated circuit discovery for mechanistic interpretability.
\newblock In \emph{Advances in Neural Information Processing Systems
  (NeurIPS)}, 2023.

\bibitem[Du et~al.(2026)Du, Gao, Zhao, Li, Wang, Lin, He, Qin, and
  Feng]{du2026}
Yanrui Du, Yibo Gao, Sendong Zhao, Jiayun Li, Haochun Wang, Qika Lin, Kai He,
  Bing Qin, and Mengling Feng.
\newblock From latent signals to reflection behavior: Tracing meta-cognitive
  activation trajectory in {R1}-style {LLM}s, 2026.

\bibitem[Elhage et~al.(2022)]{elhage2022}
Nelson Elhage et~al.
\newblock Toy models of superposition.
\newblock Transformer Circuits Thread, Anthropic, 2022.

\bibitem[G\"{a}rdenfors(2000)]{gardenfors2000}
Peter G\"{a}rdenfors.
\newblock \emph{Conceptual Spaces: The Geometry of Thought}.
\newblock MIT Press, 2000.

\bibitem[Geva et~al.(2023)]{geva2023}
Mor Geva et~al.
\newblock Dissecting recall of factual associations in auto-regressive language
  models.
\newblock In \emph{Proceedings of the 2023 Conference on Empirical Methods in
  Natural Language Processing (EMNLP)}, 2023.

\bibitem[Gould et~al.(2024)]{gould2024}
Rhys Gould et~al.
\newblock Successor heads: Recurring, interpretable attention heads in the
  wild.
\newblock In \emph{International Conference on Learning Representations
  (ICLR)}, 2024.

\bibitem[Hernandez et~al.(2024)]{hernandez2024}
Evan Hernandez et~al.
\newblock Linearity of relation decoding in transformer language models.
\newblock In \emph{International Conference on Learning Representations
  (ICLR)}, 2024.

\bibitem[Lakoff and Johnson(1980)]{lakoff1980}
George Lakoff and Mark Johnson.
\newblock \emph{Metaphors We Live By}.
\newblock University of Chicago Press, 1980.

\bibitem[Lange et~al.(2026)]{lange2026}
Georg Lange et~al.
\newblock Cross-layer transcoders are incentivized to learn unfaithful
  circuits.
\newblock LessWrong, 2026.

\bibitem[Li et~al.(2024)]{li2024}
Kenneth Li et~al.
\newblock Inference-time intervention: Eliciting truthful answers from a
  language model.
\newblock In \emph{Advances in Neural Information Processing Systems
  (NeurIPS)}, 2024.

\bibitem[Lipton(1990)]{lipton1990}
Peter Lipton.
\newblock Contrastive explanation.
\newblock \emph{Royal Institute of Philosophy Supplement}, 27:\penalty0
  247--266, 1990.

\bibitem[Ma et~al.(2026)]{ma2026}
Xuhua Ma et~al.
\newblock {CoRect}: Context-aware logit contrast for hidden state rectification
  to resolve knowledge conflicts.
\newblock \emph{arXiv preprint arXiv:2602.08221}, 2026.

\bibitem[Makelov et~al.(2024)Makelov, Lange, and Nanda]{makelov2023}
Aleksandar Makelov, Georg Lange, and Neel Nanda.
\newblock Is this the subspace you are looking for? an interpretability
  illusion for subspace activation patching.
\newblock In \emph{International Conference on Learning Representations
  (ICLR)}, 2024.

\bibitem[McGrath et~al.(2023)McGrath, Rahtz, Kram\'{a}r, Mikulik, and
  Legg]{mcgrath2023}
Thomas McGrath, Matthew Rahtz, J\'{a}nos Kram\'{a}r, Vladimir Mikulik, and
  Shane Legg.
\newblock The hydra effect: Emergent self-repair in language model
  computations.
\newblock \emph{arXiv preprint arXiv:2307.15771}, 2023.

\bibitem[Meng et~al.(2022)]{meng2022}
Kevin Meng et~al.
\newblock Locating and editing factual associations in {GPT}.
\newblock In \emph{Advances in Neural Information Processing Systems
  (NeurIPS)}, 2022.

\bibitem[nostalgebraist(2020)]{nostalgebraist2020}
nostalgebraist.
\newblock Interpreting {GPT}: The logit lens.
\newblock LessWrong, 2020.

\bibitem[Osgood et~al.(1957)Osgood, Suci, and Tannenbaum]{osgood1957}
Charles~E. Osgood, George~J. Suci, and Percy~H. Tannenbaum.
\newblock \emph{The Measurement of Meaning}.
\newblock University of Illinois Press, 1957.

\bibitem[Pal et~al.(2023)]{pal2023}
Koyena Pal et~al.
\newblock Future lens: Anticipating subsequent tokens from a single hidden
  state.
\newblock In \emph{Proceedings of the 27th Conference on Computational Natural
  Language Learning (CoNLL)}, 2023.

\bibitem[Rimsky et~al.(2024)]{rimsky2024}
Nina Rimsky et~al.
\newblock Steering {Llama} 2 via contrastive activation addition.
\newblock In \emph{Proceedings of the 62nd Annual Meeting of the Association
  for Computational Linguistics (ACL)}, 2024.

\bibitem[Sun et~al.(2026)Sun, Canziani, LeCun, and Zhu]{sun2026}
Shangwen Sun, Alfredo Canziani, Yann LeCun, and Jiachen Zhu.
\newblock The spike, the sparse and the sink: Anatomy of massive activations
  and attention sinks.
\newblock \emph{arXiv preprint arXiv:2603.05498}, 2026.

\bibitem[Templeton et~al.(2024)]{templeton2024}
Adly Templeton et~al.
\newblock Scaling monosemanticity: Extracting interpretable features from
  {Claude} 3 sonnet.
\newblock Transformer Circuits Thread, Anthropic, 2024.

\bibitem[Todd et~al.(2024)Todd, Li, Sharma, Mueller, Wallace, and
  Bau]{todd2024}
Eric Todd, Millicent~L. Li, Arnab~Sen Sharma, Aaron Mueller, Byron~C. Wallace,
  and David Bau.
\newblock Function vectors in large language models.
\newblock In \emph{International Conference on Learning Representations
  (ICLR)}, 2024.

\bibitem[Tuomi(2026)]{tuomi2026visibility}
Olli Tuomi.
\newblock A visibility threshold for top-$k$ logit-lens readouts.
\newblock Zenodo preprint, \url{https://doi.org/10.5281/zenodo.21461945}, 2026.

\bibitem[Turner et~al.(2023)Turner, Thiergart, Leech, Udell, Vazquez, Mini, and
  MacDiarmid]{turner2023}
Alexander~Matt Turner, Lisa Thiergart, Gavin Leech, David Udell, Juan~J.
  Vazquez, Ulisse Mini, and Monte MacDiarmid.
\newblock Steering language models with activation engineering.
\newblock \emph{arXiv preprint arXiv:2308.10248}, 2023.

\bibitem[van Fraassen(1980)]{vanfraassen1980}
Bas~C. van Fraassen.
\newblock \emph{The Scientific Image}.
\newblock Oxford University Press, 1980.

\bibitem[Zou et~al.(2023)]{zou2023}
Andy Zou et~al.
\newblock Representation engineering: A top-down approach to ai transparency,
  2023.

\end{thebibliography}

\end{document}